\documentclass[graybox]{svmult}

\usepackage{mathptmx}       % selects Times Roman as basic font
\usepackage{helvet}         % selects Helvetica as sans-serif font
\usepackage{courier}        % selects Courier as typewriter font
\usepackage{type1cm}        % activate if the above 3 fonts are[]
\usepackage{makeidx}         % allows index generation
\usepackage{graphicx}        % standard LaTeX graphics tool
\graphicspath{{Figures/}}
\usepackage{multicol}        % used for the two-column index
\usepackage[bottom]{footmisc}% places footnotes at page bottom
\makeindex             % used for the subject index
\graphicspath{{/}{fig/}}
\usepackage{multirow}
\usepackage{booktabs}
\usepackage{breqn}
\usepackage{tabularx}
\usepackage{amssymb}

\usepackage{caption}
\usepackage{subcaption}%for subfigures as well

\usepackage{float}
\usepackage{hyperref}

\usepackage{pgfplots}
\pgfplotsset{compat=1.7}
\usepgfplotslibrary{groupplots}

\usepackage{lipsum} % Required to insert dummy text.
\usepackage{comment}
\usepackage{todonotes}

\usepackage{soul}

\DeclareMathOperator*{\argmin}{argmin}

\DeclareMathOperator*{\mean}{mean}

\DeclareMathOperator{\match}{\stackrel{\sim}{\cup}} 
\usepackage{pifont}% http://ctan.org/pkg/pifont

\newlength\figureheight
\newlength\figurewidth
\begin{document}

\title*{Long-Term Autonomy in Forest Environment using Self-Corrective SLAM}

\titlerunning{Self-corrective SLAM\ for long-term autonomy in forest environments}

\author{Paavo Nevalainen\textsuperscript{1}*, 
  Parisa Movahedi\textsuperscript{1}, 
  Jorge Pe\~{n}a Queralta\textsuperscript{1}, \\
  Tomi Westerlund\textsuperscript{1}, 
  Jukka Heikkonen\textsuperscript{1}}
\authorrunning{Paavo Nevalainen et al.}

\institute{Paavo Nevalainen, Parisa Movahedi, Jorge Pe\~{n}a Queralta, Tomi Westerlund and Jukka Heikkonen \at \href{https://tiers.utu.fi}{Department of Computing, Faculty of Technology, University of Turku, Turku, Finland} \\\email{{ptneva, parmov, jopequ, tovewe, jukhei}@utu.fi}}

% Use the package "url.sty" to avoid problems with special characters
\maketitle

%%%%%%%%%%%%%%%%%%%%%%%%%%%%%%%%%%%%%%%%%%%%%%
%%                                          %%
%%              PAPER CONTENT               %%
%%                                          %%
%%%%%%%%%%%%%%%%%%%%%%%%%%%%%%%%%%%%%%%%%%%%%%

\abstract{
Vehicles with prolonged autonomous missions have to maintain 
environment awareness by simultaneous localization and mapping (SLAM). Closed loop correction is substituted by interpolation in rigid 
body transformation space in order to systematically reduce  
the accumulated error over different scales. The computation is
divided to an edge computed lightweight SLAM and iterative corrections
in the cloud environment. Tree locations in the forest environment 
are sent via a potentially limited communication bandwidths.
Data from a real forest site is used in the verification of 
the proposed algorithm.
The algorithm adds new iterative closest point (ICP) cases to 
the initial SLAM and measures the resulting map quality by 
the mean of the root mean squared error (RMSE) of individual 
tree clusters. Adding 4\,\% more match cases yields the mean RMSE 
0.15\,m on a large site with 180 m odometric distance. 
}


\section{Introduction}
\addcontentsline{toc}{section}{Introduction} % Adds this section to the table of contents
% \todo[inline]{Jorge, your text is superb! :D :D I'll submit the manuscript 
% at 1930.} :D Thanks

The past decade has seen a rapid evolution of methods and technologies in onboard odometry for autonomous navigation and localization. The state-of-the-art has reached a significant level of maturity in both lidar-based~\cite{shan2018lego} and visual-based odometry, among others~\cite{qin2018vins}. Nonetheless, drift in long-term operation is an inherent problem to methods based only on on-board sensors and data, with the probability of a lost position estimation increasing over time~\cite{queralta2020viouwb}. Most methods address this with loop closure~\cite{shan2018lego, qin2018vins}, where data is compared with older records if locations are repeated. In any case, long-term autonomy based only on onboard odometry data still presents significant challenges. In remote and unstructured environments such as forests, typical methods do not always apply, and loop closure can rarely be applied~\cite{li2020a}. In these scenarios, onboard processing at the edge without network-based computational offloading has inherent limitations. Challenges arise from the point of view of memory (amount of scan data to be stored for later processing), from the perspective of computational load and latency (amount of data to be used for localization through point cloud matching processes), and in terms of the update rate of the localization process (how often is the relative position computed).

Specifically, this paper deals with the problem of Simultaneous Localization and Mapping (SLAM) in unstructured forest environments with 3D laser scanners. SLAM algorithms aim at tracking the movement of the laser scanner (odometry)\ related to its surroundings and creating a composition from individual views, which consist of scanned point clouds (PC), and scanner positions and orientations. A laser scanner attached to a vehicle provides a spatial input signal which can be a very powerful component in supporting situational awareness, especially when fused with input of other sensors.

A structure from motion (SfM) study~\cite{engel2018} divides SLAM methods by two divisions: indirect and direct methods, and sparse and dense approaches. One can add three more divisions: probabilistic and non-probabilistic methods, structured and non-structured environments~\cite{li2020a}, and fixed and adaptive frame sampling. Indirect methods rely on early frame-by-frame processing, which produces a set of anchor points. The fixed frame sampling uses every frame or a fixed ratio of frames, whereas adaptive sampling tends to skip a sequence of highly similar frames. The case chosen here uses indirect anchor points from adaptively chosen frames, is non-probabilistic and is aimed to a forest environment, which is non-structured and has nearly uniformly distributed sparse anchor PC. 

\textbf{State of the art}:\ 
Iterative closest points (ICP) is the standard baseline method in SLAM. 
It is the computationally most economic choice in its simplest versions,
if the convergence can be guaranteed by the application specifics. 
If a PC is near-uniformly random, the overlap ratio of visible
cones is a good estimation for an outlier ratio $\gamma$, which is 
one of the few tuning parameters used by robust ICPs~\cite{chetverikov2002}. 
The outlier ratio $\gamma$ limits the ICP matching process to 
$1-\gamma$ part of the match pairs and reduces the accumulation of the 
odometric error. If the matches occur in a geometrically consistent 
zone between two PCs, the match overlap ratio $\lambda$ can be estimated
by $\lambda\approx 1-\gamma$.

The strategy of forcing a minimum overlap does not guarantee global convergence, though. 
A globally optimal method with a proven convergence is 
Go-ICP~\cite{yang2015}, which can detect a difference between a local and global ICP 
match in the case of near-uniform PC. Other 'global' methods try to smooth 
the mean matching error, which is the target function, by various means,
but fail to guarantee the convergence to a global optimum. 
   
Near uniform randomness gives a chance to tighten the conditions of 
branch-and-bound (BnB)\ limit estimates used in~\cite{yang2015}. 
We use Go-ICP as a backbone of a naive and risky SLAM method 
\textit{pcregistericp}()~\cite{matlab2020} since there are no SLAM\ methods 
specifically suited for sparse uniformly random PCs 
to the best of our knowledge.

When dealing with a limited view cone (less than 360$^o$ view), a problem
similar to closed loop detection~\cite{williams2009,wang2019} occurs 
each time the view cone coincides with a much older frame. This happens
in a small scale of 2 to 5\,m but it can also happen over distances of
0.5\,km to 1\,km. 

\textbf{Motivation}:\  
Autonomous mobile robots and specifically unmanned aerial vehicles (UAVs) have seen an increase penetration for forest surveying and remote sensing~\cite{chisholm2013uav, sankey2017uav}. Owing to the unstructured environments that forests represent, autonomous navigation presents inherent challenges. Key issues appear in the areas of localization and mapping, where one has to take into account several key points in a local scope to make the SLAM computationally feasible~\cite{li2020a}. Moreover, an autonomy stack for forest navigation ought to consider long-term autonomy (e.g., owing to the long distances that UAVs can traverse over long times, or the longer time that ground vehicles can operate). Typical odometry techniques relying on loop closure do not suffice because locations are not repeated often. In particular, we are interested in lidar-based odometry, localization and mapping with methods that can be used for both unmanned ground vehicles (UGVs) and UAVs.

Taking into account these considerations, there is a need for more advanced techniques for long-term autonomy exploiting registrations of the same objects even from distant locations. This approach differs from traditional loop closure as there can be 
several partial overlaps of frames over different time scales, and the partial overlap 
may occur over large distances. Modern laser scanners, even low-cost solid state LiDARs, are capable of measuring distance to objects up to several hundreds of meters~\cite{ortiz2019initial}.  

In addition to the accuracy of localization over long distances, the majority of the state-of-the-art lidar-based SLAM algorithms require relatively high computational resources to operate in real-time~\cite{shan2018lego, lin2020loam}. Moreover, the amount of points in a single scan PC have increased to millions per second in recent years. Solid state lidars with limited field of view (FoV) are only able to detect a reduced number of features in a single scan, but the scan density increases significantly. Therefore, techniques that need to compare all points (e.g., ICP) or traditional feature extraction techniques do not scale well. From the perspective of map-based localization, a similar issue arises with approaches such as the normal distribution transform (NDT)~~\cite{einhorn2015generic}.

Our approach extends the idea of loop closure to track features (i.e., tree stems) over long-term autonomous operation, which can be stored and compared in the form of sparse PCs. Relying in sparse PCs that contain only uniformly distributed features, we are able to both reduce the processing time for localization (in terms of point cloud matching) and the size in memory of larger-scale maps.

In general, we see a gap in the literature in approaches to long-term autonomy and self-corrective localization leveraging the matching of uniformly random feature points. To the best of our knowledge, this together with exploiting sparse PCs for faster processing in unstructured environments has not been addressed yet. Moreover, our approach can be leveraged for managing and accounting for the actively rotating view cone of modern solid-state lidars.

Recently, there have been progress in rigid body interpolation~\cite{lynch2017}
mainly applied in the robotics field. There is an advantage in having 
both rotation and translation addressed at the same time. To our 
understanding, this provides a chance to address the odometric 
consistency independently of the scale of the localization problem. 

\textbf{Contribution}: 

We propose a method to reduce the cumulative match error
by adding extra ICP matches, which comprise a large
time interval. Frames within the interval are squeezed together 
by an interpolation scheme reducing the imprint of sets of 
anchor points associated together in the final map. 
The process reduces the noise and blur in the final map,
increases the odometric accuracy and solves both small scale
closed loop occurrences due to the work cycle movements, 
and large scale closed loop problems.

In summary, in this paper we address the following three issues: (i) a self-corrective localization algorithm able to incrementally increase the accuracy of the produced environment map without relying on loop closure; (ii) memory efficiency and computation at the edge by relying on sparse point clouds and long-term tracking of features; and (iii) an adaptive approach that adjusts the positioning update rate based on the available data at a given time.

%------------------------------------------------

\section{Methods}
\begin{figure}[h!]\centering 
\includegraphics[width=6cm]{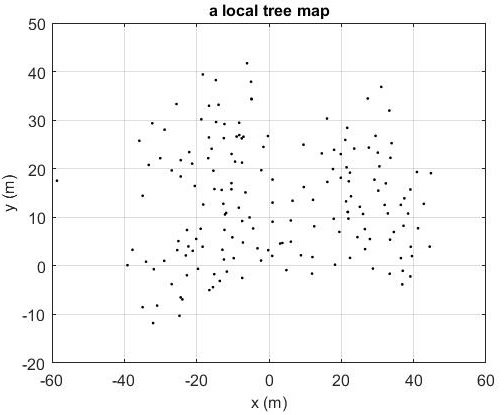}
\includegraphics[width=5.5cm]{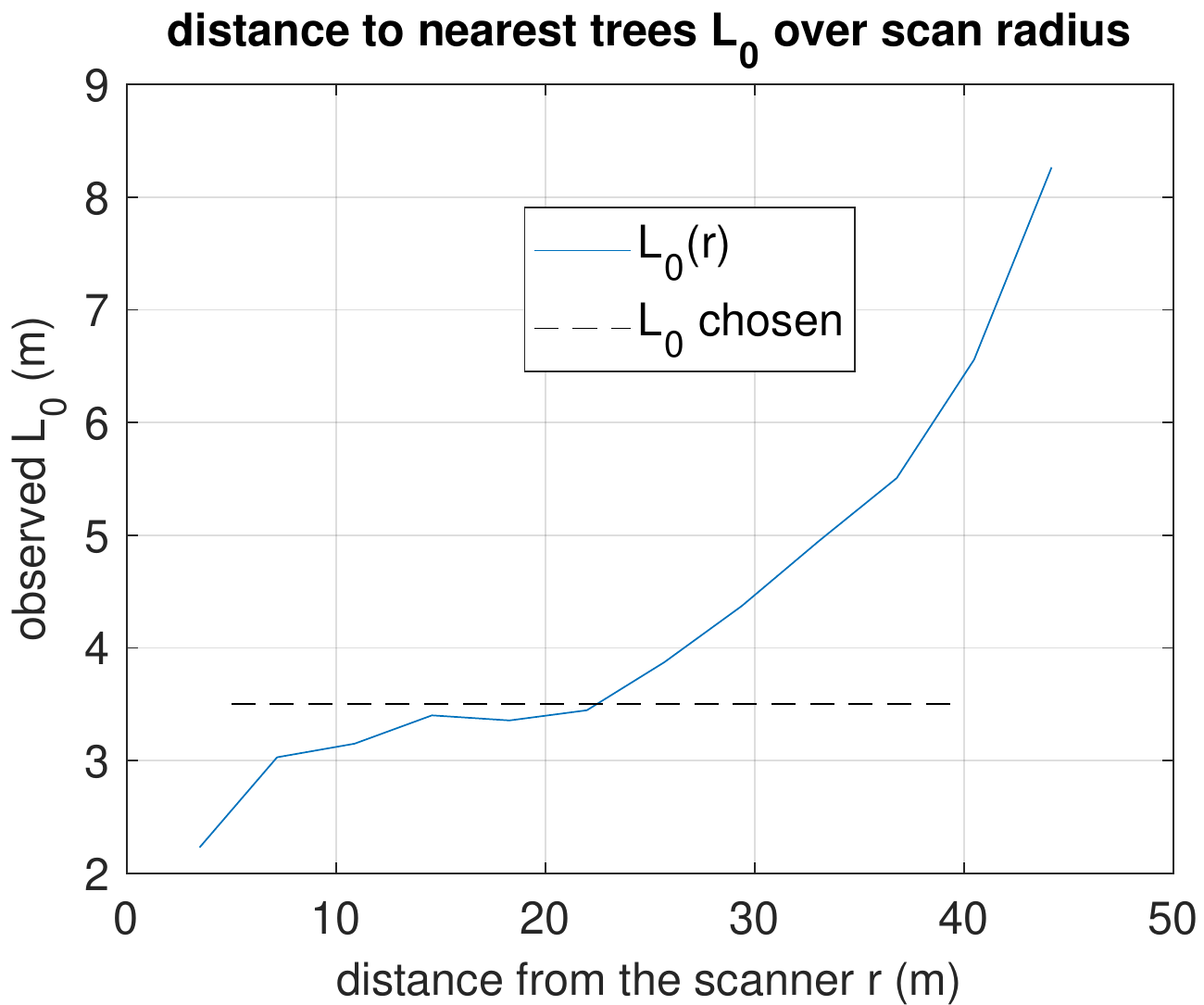}
\caption{Left: A single scanner view. Each 
point represents an edge-computed registration 
of a tree. Points are 3D even shown in a horizontal
projection. Right: Mean distances to nearest trees
at each scanner range. $L_0=3.5$ m is over all 
the trees and is depicted with a dash line.}
\label{fig:aFrame}
\end{figure}

\textbf{The site and the data}: The test data is from 
a forest operation site in Pankakangas at Lieksa, Eastern
Finland (63$^o$ 19.08' N, 30$^o$ 11.57' E). The data was recorded on 
August 2017 in co-operation with participants enlisted in the 
Acknowledgements section. The sample has a strip road 
of length 130 m. There are 9009 data frames and 3.7 GB of \textit{.pcap} 
data. The total number of trees is 680 and one frame includes 130 trees
on the average. An example scanner view is depicted in the left 
detail of Fig.~\ref{fig:aFrame}. The mean distance to nearest trees 
is $L_0= 3.5$ m excluding the peripheral zone at a distance of 40 m.
The right detail shows how the mean distance increases over the radial
distance.

\textbf{Methodology}: 
A simple and fast ICP method \textit{pcregistericp}()~\cite{chen1992,besl1992} 
implemented in Matlab~\cite{matlab2020} produces rigid body transformations,
which can be used to build an environment map from sparse key points, each 
key point representing a detected tree in a scanner view. The problem is
to improve this rather low-quality tree map by selecting a small 
set of promising pairs of frames and producing a computationally more 
expensive and more accurate match using Go-ICP~\cite{yang2015}. Each
extra match between frames $j$ and $i$ may engulf several frames, which 
should be properly adapted to the newly introduced and very reliable 
match. The exponential interpolation of rigid body transformations is
used for that purpose.

We introduce first the rigid body transformation as a homogenous operator and its 
logarithm and exponentiation, which are the core of the interpolation method. 
A novel aspect is the operator power being in the matrix form. We show that 
the interpolation is contractive meaning that the final PC\ map improves 
(individual tree clusters become sharper) per each addition of extra matches. 
The sharpening can be measured internally by minimization 
of the mean match error and externally by observing the mean radius of clusters in the 
final map. Finally we define the control parameters for 
the branch-and-bound (BnB) of the global minimum search of the SLAM match.

\subsection{Operator exponentiation}\label{sec:exponentiation}
A rigid body transformation $\tau\in SE(3)$ in the special Euclidean group
$SE(3)$ consists of one rotation represented by a rotation 
$R\in SO(3)$ within the special orthogonal group $SO(3)$  
followed by and a translation $p\in\mathbb{R}^3$. The treatise 
uses the notation and conventions of~\cite{lynch2017}.
The transformation $\tau$ can be seen as a homogenous
mapping $\tau:\ q\mapsto q'$: 
\begin{equation}
   \begin{pmatrix}
     q' \\ 1
   \end{pmatrix} = \overbrace{
   \begin{pmatrix}
     [R] & p\\ 0 &\ 1
   \end{pmatrix}   }^{[\tau]}
   \begin{pmatrix}
     q \\ 1
   \end{pmatrix},
   \label{eq:tau}
\end{equation}
where braces $[.]$ depict the matrix representation of an operator. 
Transformation $\tau(R,p)$ is defined by a pair of a rotation and 
a translation. An alternative parameterization is  $\tau(\vec\omega,\theta,v)$
where a unit axis $\vec\omega=(\omega_1\,\omega_2\,\omega_3)\in\mathbb{R}^3$ is the rotation axis, $\theta$
is a rotation angle around that axis, and $v\in\mathbb{R}^3$ is the tangential
direction, where the origo moves in the beginning of that rotation. 
Note that $\vec\omega$ is the unit eigenvector of $[R]$ associated 
to the eigenvalue 1:  $[R]\vec\omega= \vec\omega$.

A twist $S([\omega],v)$ combines two of the elements of a rigid body transformation,
and its matrix form is:
\begin{equation}
 [S] = \begin{pmatrix}
     [\omega] & v\\ 0 &\ 0
   \end{pmatrix},
   \label{eq:S}
\end{equation}
where  $[\omega] a=\vec\omega\times a$ for any $a\in\mathbb{R}$, or, as written open in the matrix form:
\begin{equation}
[\omega] =
   \begin{pmatrix}
             0 & -\omega_3 & \omega_2 \\ 
      \omega_3 &         0 & -\omega_1 \\
     -\omega_2 &  \omega_1 & 0
   \end{pmatrix}.
   \label{eq:omegaSkewAssembly} 
\end{equation}
Note that this definition gives $[\omega]$ a cyclic property: $[\omega]^3=-[\omega]$
which will be used, when dealing with series expansions of $e^x$, $\sin{x}$
and $\cos{ x}$. Exponentiation of $[S]\theta$ gives us:
\begin{equation}
 e^{[S]\theta} = \begin{pmatrix}
     e^{[\omega]\theta} & G(\theta)v\\ 0 &\ 1
   \end{pmatrix}=[\tau],
   \label{eq:eS}
\end{equation}
where the right equality can be settled by setting $v=G^{-1}(\theta)p$ and 
$e^{[\omega]\theta}=[R]$. The twist gain function $G(\theta)$ unfolds by 
the exponentiation series and the rotation matrix term can be expanded to a closed form:
\begin{equation}
   [R]=I+\sin{\theta}[\omega] + (1-\cos{\theta})[\omega]^2,
   \label{eq:R}
\end{equation}
which is the well-known Rodriguez formula. Finally, raising $\tau$ to a power $u\in\mathbb{R}$, one gets:
\begin{equation}
 [\tau^u] = \begin{pmatrix}
     [R^u] & p_u\\ 0 &\ 1
   \end{pmatrix},
   \label{eq:tauU}
\end{equation}   
where 
\begin{eqnarray}
  [R^u]&=&I+\sin{\theta u}[\omega] + (1-\cos{\theta u})[\omega]^2\label{eq:Ru}\\
  p_u  &=&G(\theta u)G^{-1}(\theta)p. \label{eq:pu}
\end{eqnarray}

The homogenous representation allowed a definition of a matrix power of a rigid body transformation limited to $SE(3)$. To signify this limitation, we write $[\tau^u]$ and not $[\tau]^u$, since
the wide realm of general matrices is perilous~\cite{moler2003}, what comes
to exponentiation and taking logarithms. The same argument holds to notation
with $[R^u]$.

The twist gain function $G(\theta )$ is opened next:
\begin{equation}
   G(\theta)= I\theta+ (1-\cos{\theta})[\omega]+ 
      (\theta- \sin{\theta})[\omega]^2.
      \label{eq:G}
\end{equation}
Its inverse is needed in the Equation~\ref{eq:pu}: 
\begin{equation}
   G^{-1}(\theta)= \frac{1}{\theta}I- \frac{1}{2}[\omega]+ 
      \left(\frac{1}{\theta} - \frac{1}{2\tan(\theta/2)}\right)[\omega]^2 .
      \label{eq:invG}
\end{equation}

One can easily see that there is a singularity in $G^{-1}(\theta)$ when 
$\theta\rightarrow 0$. But the product in Eq.~\ref{eq:pu} stays defined,
albeit it needs a Taylor series expansion\footnote{...or a min-max polynomial definition, 
which is excluded from this treatment.} at $\theta=0$. This is needed because 
the homogenous formulation chosen here is not a conformal theory~\cite{dorst2009}.
The product $G(\theta u)G^{-1}(\theta)$ develops to:
\begin{multline}
  G(\theta u)G^{-1}(\theta)=Iu+\left[A(\theta,u)-\frac{\sin{\theta u}}{2}\right][\omega]+\\
  +\left[u-\frac{1-\cos{\theta u}}{2}-\ B(\theta,u)\right][\omega]^2,
  \label{eq:GG}
\end{multline}
where:
\begin{equation}
   A(\theta,u)= \frac{1-\cos{\theta u}}{2\tan{\theta/2}},\; 
   B(\theta,u)= \frac{\sin{\theta u}}{2\tan{\theta/2}},
   \label{eq:AandB}
\end{equation}
which are both of a form 0/0 at $\theta=0$.
Taylor series developed at $\theta=0$ give:
\begin{eqnarray}
   A(\theta,u)&= \frac{u^2}{2}\theta  - (u^2+u^4)\theta^3/24 + \mathcal{O}(\theta^4)
   \label{eq:A}\\
   B(\theta,u)&= u - (u/12+u^3/6)\theta^2+ \mathcal{O}(\theta^4).
   \label{eq:B}
\end{eqnarray}
 
Note that small values of $\theta$ will often occur with the intended application,
whereas large values $\theta\approx\pi$ occur seldomlu, if ever.

\begin{figure}[h!]\centering 
\includegraphics[width=8cm]{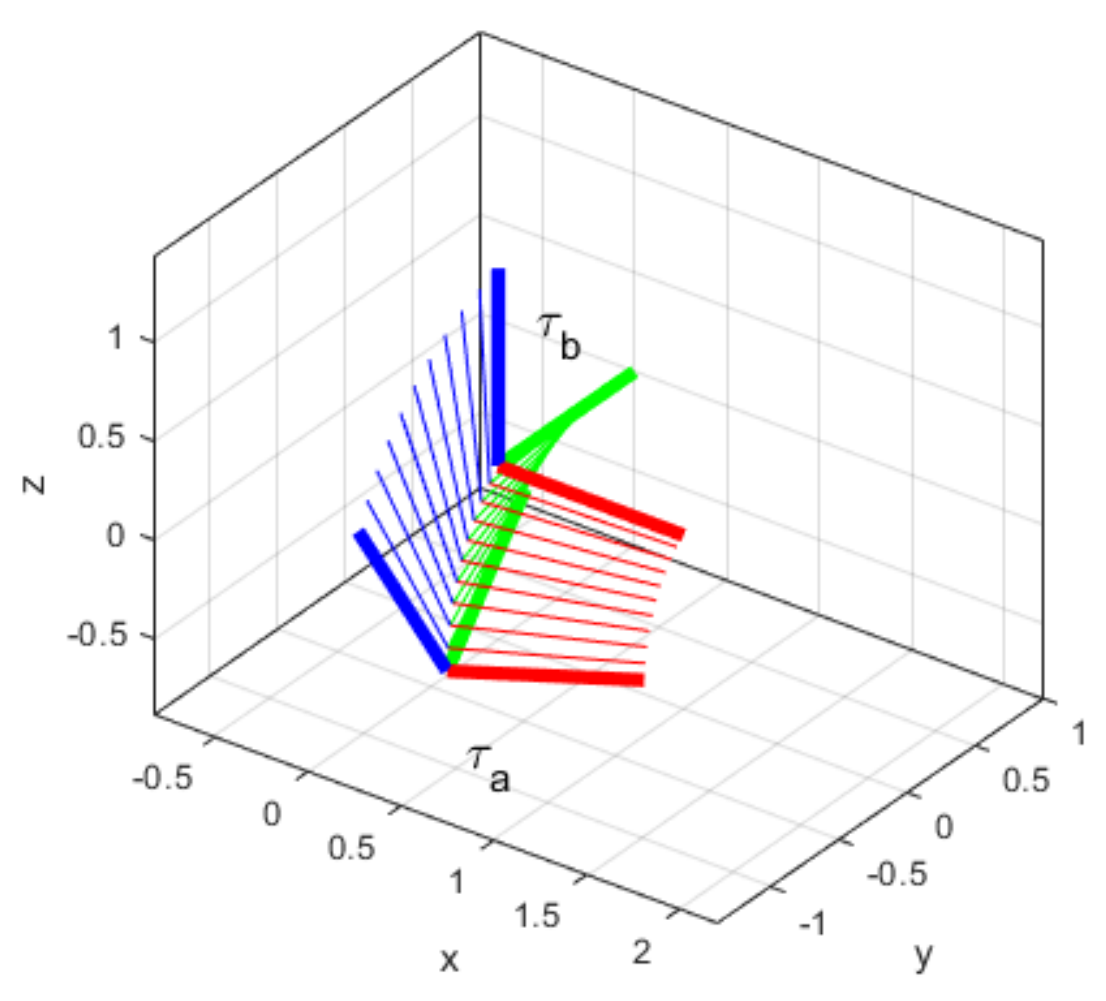}
\caption{Example of a power interpolation $\tau_u=\tau_a^{1-u}\tau_b^u$.
When $u=0\rightarrow 1$, $\tau_u=\tau_a\rightarrow \tau_b$
smoothly between two rigid body frames $\tau_a$ and $\tau_b$. 
Columns of $[R]$ form the orthogonal frame axes, which are shown in 
red, green and blue.}
\label{fig:aSkew}
\end{figure}

Extraction of $[\omega]$ and $\theta$ from a given $[R]$ is called 
taking a rotation logarithm, since $[R]=e^{[\omega]\theta}$.
The exact logarithm algorithm is given in~\cite{lynch2017} and has 
two special cases for $\theta\approx 0$ and $\theta\approx\pi$. The intended application of 
the transformation matrix power is such that one needs to solve the Eq.~\ref{eq:tauU} several times 
with different values of $u$, so the constant parts $[\omega]$ and $\theta$ 
are pivotal. Fig.~\ref{fig:aSkew} shows an example, where a rigid body
co-ordinate frame $\tau_a$ is interpolated to another frame $\tau_b$ using
11 values $u\in[0,1]\mathbb{R}$.

The matrix power in Eq.~\ref{eq:tauU} has one special case of pure translation
where there is no rotation ($\theta=0$ and $[R]=I$) with:
\begin{equation}
\begin{pmatrix}
     I & p\\ 0 &\ 1
   \end{pmatrix}^u =
   \begin{pmatrix}
     I & pu\\ 0 &\ 1
   \end{pmatrix}.
   \label{eq:specialCase} 
\end{equation}
Naturally, this special case should be covered by a general solution 
of the vector $p_u$.  
As a sanity check, setting $\theta\approx 0$ leads to $p_u\approx pu$ 
for all $u\in\mathbb{R}$.
By setting $u=1$ and after a tedious trigonometric manipulation, one 
gets $p_u=p$ for all $\theta$. Although 
the Eqs.~\ref{eq:GG}-~\ref{eq:B} are novel in the context 
of the matrix power of the homogenous formulation, a similar 
Taylor series approach has been presented for dual quaternion
exponentiation and logarithm in~\cite{dantam2018}, and one could 
construct a similar transformation power $\tau^u$ using
suitable quaternion libraries. The value of the Eqs.~\ref{eq:GG}-~\ref{eq:B}
is that the odometric process described in the next section can proceed within a usual matrix infrastructure. The computational price tag of two alternative
formulations (dual quaternions and homogenous co-ordinates) for taking 
multiple matrix powers is surprisingly close to each other for large PCs. 
%We leave a proper analysis out of this presentation, though.

\begin{figure}[h!]\centering 
% Using \begin{figure*} makes the figure take up the entire width of the page
\includegraphics[width=8cm]{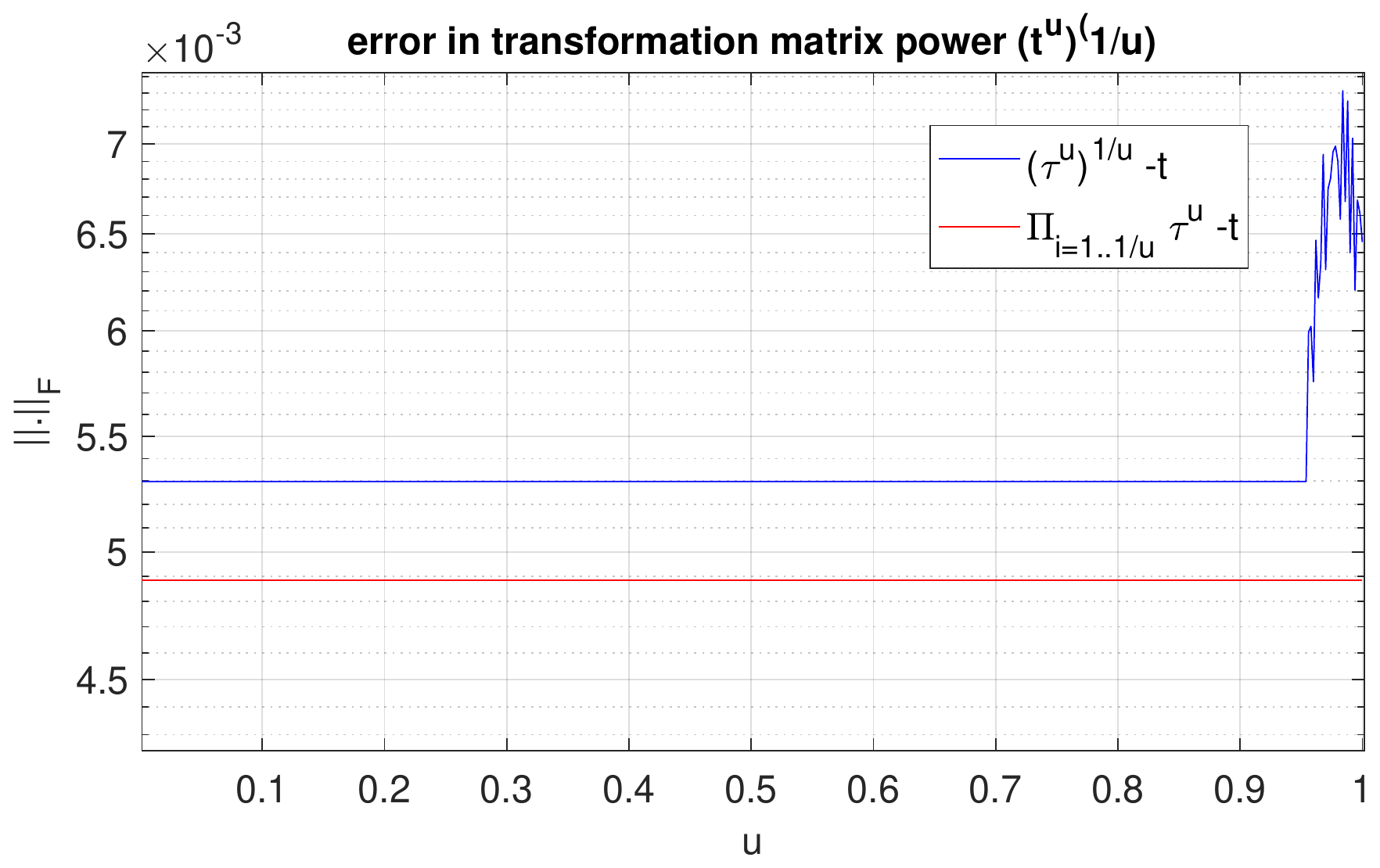}
\caption{A numerical verification of rising rigid body transformations 
to a power $u$.
Blue line depicts taking  a power  $u$ and  then $1/u$. Red line is for 
taking a power $u$ followed by a number ($1/u\in\mathbb{N}$) of matrix 
multiplications.}
\label{fig:TuError}
\end{figure}

A study on the error of the exponentiation of the rigid body transformation
follows. Two consecutive operations $\|[(t^u)^{1/u}]-[t]\|$ produce an 
error shown in the Fig.~\ref{fig:TuError}. The result is the average over
300 rigid body transformations with a uniform distributed $p\in[-1,1]^3\subset\mathbb{R}^3$
and $R(\vec\omega,\theta)$, where $\vec\omega\in S^2$ is uniformly distributed
over a unit sphere $S^2\subset\mathbb{R}^3$ and $\theta\in[0,\pi)$ is also uniformly distributed. The sequential matrix multiplication version with $1/u\in\mathbb{N}$ has been
provided (red line) alongside the usual $u\in[0,1]\subset\mathbb{R}$ 
matrix power test (blue line).
As can be seen from the Fig.~\ref{fig:TuError}, the computational accuracy 
is not a problem near $u=0$, and $u\approx 1$ does not usually occur. The overall 
accuracy in the multiplication case is $4.8\times 10^{-3}$, which
is enough for practical implementation. Both tests are clearly conservative
when compared to actual computations, which generate new operators $\tau^u$ 
from constant values $\theta$ and $[\omega]$.
 
\subsection{Rigid body motion interpolation}\label{sec:skew}
This Section expands the presentation in~\cite{nevalainen2020a}
and uses the notation of~\cite{lynch2017}. 
The detailed definitions are provided since the formulations come 
from a variety of sources.
Odometry is built by matching  sequential PCs $P_l$ and $P_{l+1}$
in the coordinate system of the frame $l$ by estimating a transformation $\tau_{l+1\,l}$ (from a frame $l+1$ to the frame $l$).
A simple and fast ICP method \textit{pcregistericp}()
of Matlab~\cite{matlab2020,chen1992,besl1992} is applied
to produce a sequence of rigid body transformations 
$\tau_{l+1\,l}$ from a frame $l+1$ to a frame $l$. This process
is not secure, it is possible to have an erroneous match, which is 
off some 2-10 meters.

The combination of two PCs achieved by a succesful match is denoted by $.\match.$ as:\ 
$P_l \match P_{l+1}t_{l+1\,l}$, where $t_{ij}=[\tau_{ij}]^T$ because,
unlike in the definition of Eq~\ref{eq:tau}, points are now columns of 
a PC matrix.
A match between frames $l+1$ and $l$ includes inaccuracies $e_l$, and 
identification of outliers (points not matched to any point) and of matching 
pairs of points. The final SLAM result has all frames matched to the first frame:
\begin{equation}
  P=\match_{l=1}^n P_l\,t_{l1} \label{eq:globalization}
\end{equation}
where $n$ is the number of frames, $t_{11}=I$ and total transformation matrices
$t_{l1}$ are built iteratively from local matches 
$t_{l+1\,l}$ by:\ $t_{l+1\,1}=t_{l1}t_{l+1\,l}$.

The PCs $P_l\,t_{l1}$ each contribute to the total map $P$.
A typical ICP process produces a chain of stepwise transformations 
$\tau_{l+1}^l,\,1\le l<n$. 
One can recover a random transformation between frames $i$ and $j$ 
from total transformations by:
\begin{equation}
    t_{ij}= {t_{j1}}^{-1}t_{i1}.
    \label{eq:tij}
\end{equation}

The iterative application of the Eq.~\ref{eq:globalization} is called 
globalization (or SLAM process). The odometry problem is solved when 
the globalized translation vectors are extracted. The path $Q$ of 
the vehicle is:
\begin{equation}
   Q= \{q_{i1}\,|\, \tau_{i1}= (R_{i1},\,q_{i1})\}_{i=1...n}
\end{equation}

\begin{figure}[h!]\centering 
\includegraphics[width=8cm]{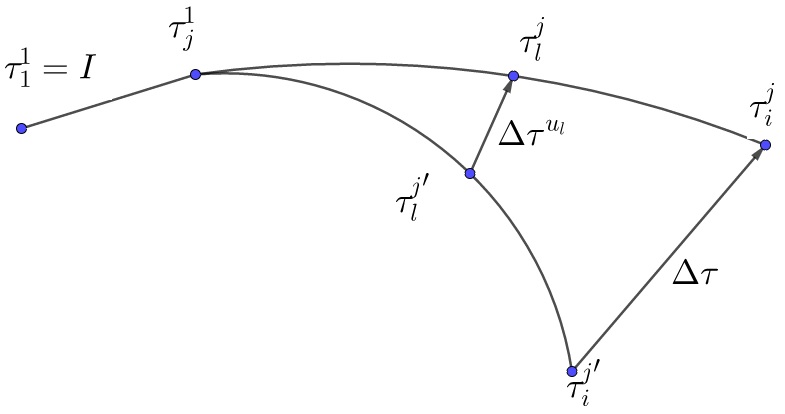}
\caption{An update from  $\tau_{i1}$ to ${\tau_{i1}}'$ makes the old 
path $j+1,...,i-1$ incompatible. It can be corrected by any interpolation
 scheme, e.g. the Eq.~\ref{eq:tauPropagation1}.}
\label{fig:tauCorrection}
\end{figure}

The basic scenario of the self-corrrective odometry is depicted in Fig.~\ref{fig:tauCorrection}
using two paths $Q$\ and $Q'$ to represent a situation, where some of the transformations
$t_{l+1\,l},\, j\le l < i$  have been judged inaccurate, 
noisy or inexact by some criteria. The criteria is usually related to the blurriness of the global map. Then a corrective check is being performed from the frame $j$ to the frame $i$ producing an improvement of a match.
Formally, an error measure 
$e(P_j\match P_i\,{t_{ij}}')< e(P_j\match P_i\,{t_{ij}})$
of a match improves, when a new match ${t_{ij}}'$ is used instead of 
a synthetized transformation $t_{ij}$ of Eq~\ref{eq:tij}. 
Now, all the intermediary PCs $PC_l,\,j<l<i$ need to be updated. 

A solution is to force the differences $\Delta\tau_l$ to have boundary
conditions $\Delta t_j=I$ and $\Delta t_i=\Delta t={t_{ij}}^{-1}{t_{ij}}'$.
A rather obvious interpolator is by $\Delta t_l=\Delta t^{u_l}$, where
$0\le u_l\le 1$ with obvious end conditions $u_j=0,\, u_i=1$.
Assuming representative powers $u_l$ are defined for each transformation
$t_{lj},\,j\le l\le i$, one can solve new values ${t_{lj}}'$:
\begin{equation}
    t_{l1}:= {t_{l1}}'= t_{j1} t_{lj}\Delta t^{u_l},
    \label{eq:tauPropagation1}
\end{equation}
where ':=' denotes a computational substitution of a new value.
As a sanity test, by setting $l=j,\,u_l=0$ one gets:
${t_{j1}}'= t_{j1}I^2= t_{j1}$. And by setting $l=i,\,u_l=1$ one gets:\\
${t_{i1}}'= t_{j1} t_{ij}{t_{ij}}^{-1}{t_{ij}}'= t_{j1}{t_{ij}}'$.
The path following after the frame $i$ changes after this update, too. 
The rest of the frames have to be corrected to align properly with 
the updated value ${t_{i1}}'$:
\begin{equation}
   t_{k1}:={t_{i1}}'\overbrace{{t_{i1}}^{-1}t_{k1}}^{t_{ki}}
    \label{eq:tauPropagation2}
\end{equation}

One question remains:\ how to choose the power $u_l$, given a SLAM history 
$\{t_{l1}\}_{j\le l\le i}$? There are several possibilities but for 
numerical experiments we used the simplest possible strategy, the relative 
continuous index:
\begin{equation}
   u_{l}=(l-j)/(i-j),\; j\le l \le i.
   \label{eq:ulRule}
\end{equation}

\textbf{Contractive property}: We propagate change on odometric path 
$j,...,i$ by using $[\tau^u]$ as a correction term. 
As long as all the involved powers $u_l$ are confined to the unit 
interval $0\le u_l\le 1$, the new PC\ mappings 
$\match_{l=1}^i P_l{t_{l1}}'$ contract, i.e. all the involved
rotations ${\theta_l}'$ for each sub-match $l$ of a corrective step 
$\tau_i^j$ become smaller ${\theta_l}'\le\theta_l$ and the magnitude of 
translations ${q_l}'$ gets reduced ${q_l}'\le q_l$. The proof is based 
on the monotonicity of terms $G(\theta u)G^{-1}(\theta)$ (see Eq.~\ref{eq:pu}) 
and $\theta u$ on the basis of $\{I,[\omega],[\omega]^2\}$.
A visual evidence of this is shown in~\cite{nevalainen2020a}, 
where tree clusters get less dispersed on each step of iterative
improvement.

\subsection{Branch-and-bound limits}\label{sec:bnb}
The globally convergent ICP method Go-ICP~\cite{yang2015}
uses two coefficients $\sigma_t$ and $\sigma_r$ to set
up the granularity of the transition and rotation search space in 
the BnB search grid, respectively. Two coefficients $\gamma_t=\sqrt{3}\sigma_t$ 
and $\gamma_{rp}=2\sin(\min(\sqrt{3}\sigma_r/2,\pi/2))\|p\|$ 
define the local lower bound of the minimum match error at
a point $p\in P$ in the original scanning frame. Note that
rotational term $\gamma_{rp}$ indeed depends on the point $p$.
A minimum bound $\underline{e}_{ij}$ of a match error $e_{ij}$ 
between frames $i$ and $j$ is:
\begin{equation}
  \underline{e}_{ij}^2=
     \sum_{p\in P}\left(\max(e_p-\gamma_{rp}-\gamma_t,0)\right)^2.
  \label{eq:eMin}
\end{equation}

Tree locations in 
a forest usually have a nearly uniform distribution, which
can be described by a mean distance $L_0$ m between natural 
neighbors (a concept defined in the next paragraph). 
The right detail of Fig.~\ref{fig:aFrame} depicts
the $L_0$ distribution over the scanning range $r$, and even 
the point density depends on range, the zone $10\le r \le 26$ m
with $3.0 \le L_0 \le 4.0$ m is large and populated enough for 
our purposes. In our data samples, $L_0= 3.5$ m was the observed
average specific to the data collection site.

To make the definition of $L_0$ more formal, we define
natural neighbors $q\in N(p)\subset P$ of a point $p$ as those
points, which get connected by an edge $(p,q)\in E\subset P^2$
in a Delaunay triangularization $(P,E,T)$ of a point set $P$. 
There, $E$ are edges of Delaunay triangles $T\subset P^3$.
Then, the mean distance is:
\begin{equation}
   L_0= \mean_{p\in P}\mean_{q\in N(p)} \|p-q\|. \label{eq:delta0}
\end{equation}
If a magnitude $\delta=\|q\|$ of a pure translation from a perfect match is 
smaller than $L_0/2$, $\delta< L_0/2$, the ICP convergence is very likely. 
This will be shown later by a numerical experiment. We define this limit 
as $\delta_0$:
\begin{equation}
    \delta_0= L_0/2 \label{eq:delta0b}.
\end{equation}
For reference; a hexagonal lattice is the optimal packing on points and having 
two such PCs switched randomly produces a mean match error $e= 0.35 L_0$.

\begin{figure}[h!]\centering 
\includegraphics[width=4.5cm]{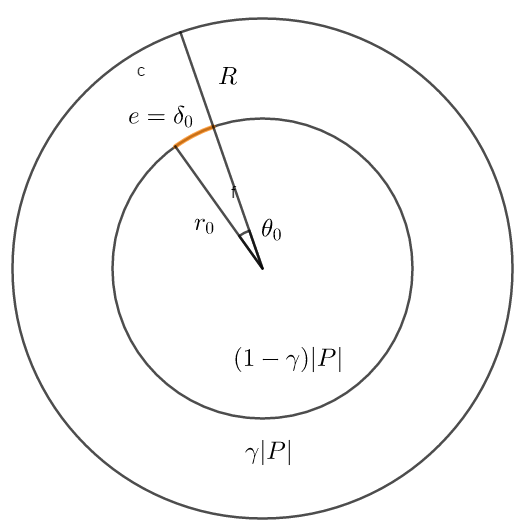}
\caption{A rotation $\theta_0$ produces a mismatch $\delta_0$ 
to $\gamma|P|$ point pairs starting from $r_0$, when a uniform 
distributed PC is rotated around its center.}
\label{fig:thetaEffects}
\end{figure}

Other important parameters characterizing the scanned PCs are the scanning 
scope $R$ and the allowed outlier ratio $\gamma$, which makes the standard 
ICP method somewhat more robust. Outliers are points without a proper match. 
Fig.~\ref{fig:thetaEffects} shows a circular PC being rotated by an angle
$\theta_0$. At a distance $r_0$, $\delta_0=r_0\theta_0$. The radius $r_0$
divides the disc to two parts with ratios $\gamma\,:\,1-\gamma$. A simplfying
assumption is being made that all the outer point pairs do not match, and 
all the inner point pairs do match, so that $1-\gamma=\pi r_0^2/(\pi R^2)$
and one can solve $\theta_0$:
\begin{equation}
    \theta_0= \frac{\delta_0}{\sqrt{1-\gamma}R}.
    \label{eq:theta0}
\end{equation}

A justification for this simple derivation of the limit $\theta_0$ is 
that a rotation with $\theta_0 <\theta$ of two identical uniformly distributed 
PCs  produces a situation similar to a case of two i.i.d PCs at the outer 
zone $r_0<r$. 

A simple ICP is assumed to succeed when: 
\begin{equation}
    \left(\frac{\delta}{\delta_0}\right)^2+
    \left(\frac{\theta}{\theta_0}\right)^2 \le 1,
    \label{eq:convergenceTest}
\end{equation}
where $\delta$ is the known magnitude $\delta=\|q\|$ of 
the known match $\tau([\omega],\theta,q)$ and $\theta$ is 
the known horizontal rotation from the correct match. 
This means that for a possible grid search or BnB approach, 
the two parameters $\lambda_0$ and $\theta_0$ define the grid granularity.

A complete misalignment has mean error $e$ between pairs of 
matching points $e\approx\delta_0$, and a complete alignment 
equals the registration noise $\epsilon$: $e\approx\epsilon$.
The ICP match succeeds at the limit $\theta=\theta_0$ since 
the point density decreases, and the local $L_0$ increases, when 
scan radius $r$ grows, see the right detail of Fig.~\ref{fig:aFrame}. 

The BnB search grid can be set to a granularity, where the final
attempt at the finest level of the search hierarchy can be 
safely done by a simple ICP. By this arrangement the BnB grid
does not need to extend to actual tolerances sought after. 
Numerical values in our example data are $L_0=3.5$ m, 
$\gamma= 0.6$, $\delta_0=1.75$ m and $R=35$ yielding 
$\theta_0= 3.7^{o}$.

\begin{figure}[h!]\centering 
\includegraphics[width=9cm]{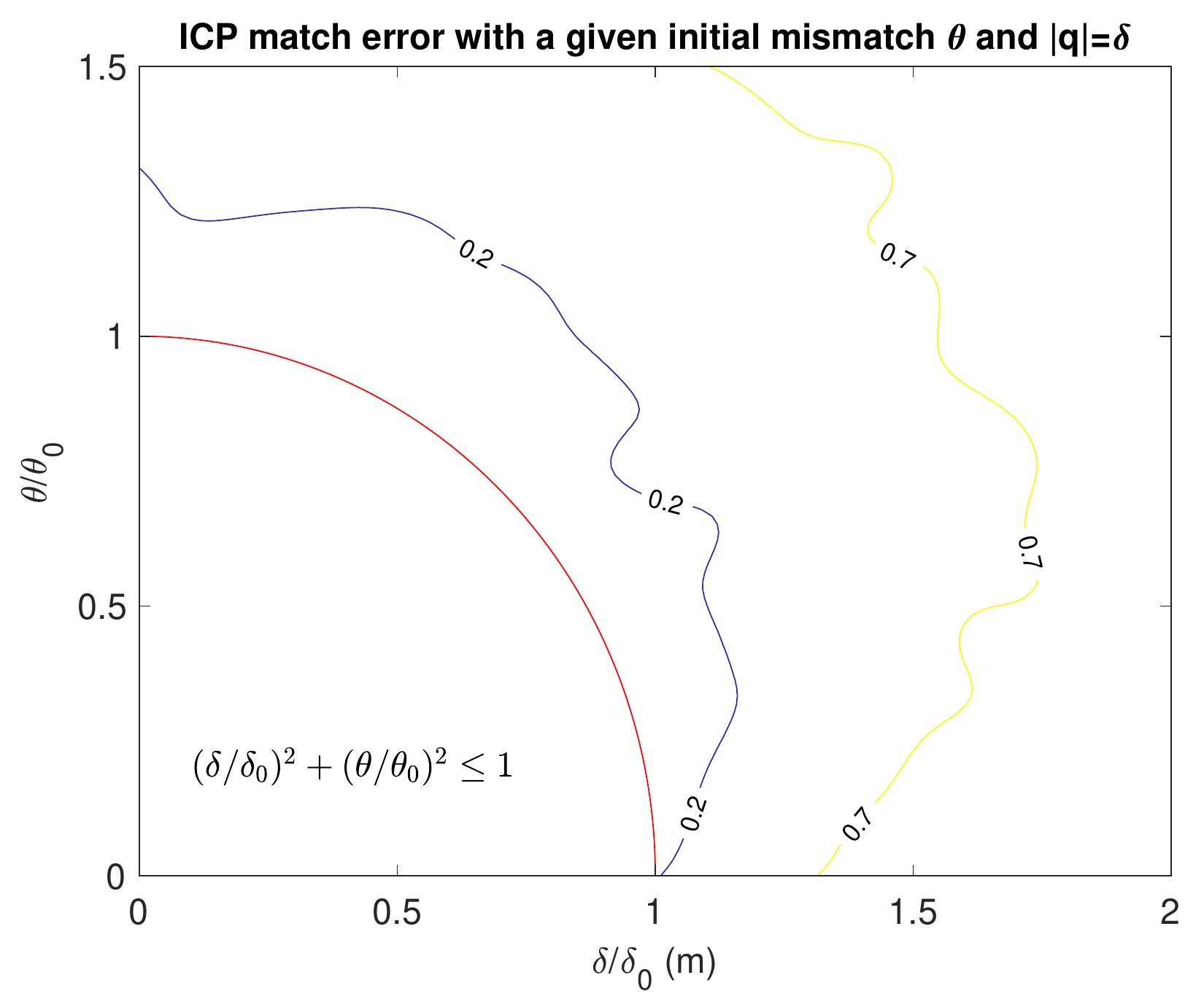}
\caption{Match error $e$ (m) of a simple ICP method, when initial 
 mismatch state $\tau(\theta,\delta)$ is known. The area 
 inside the circle is where the convergence is quaranteed 
 for uniform distributed PCs with $L_0=2\delta_0$.} 
\label{fig:convergenceCondition}
\end{figure}

The convergence condition of Eq.~\ref{eq:convergenceTest} requires
a verification. Fig.~\ref{fig:convergenceCondition} 
summarizes a test setting, where matched PC pairs drawn from the data 
were artificially separated by $\tau(\theta,\delta q^0)$, where translations 
were taken to several directions encoded by unit vectors $q^0$ 
to register the effect over the translation range $\delta$.
Two contour lines with the match error $e= 0.2$ m and 
$e= 0.7$ m are shown. A successful match has $e\le 0.2$ mm 
since this does contribute well to the desired final tree map accuracy.
The point registration noise $\epsilon$ from the tree registration 
process is approximately $0.05 < \epsilon < 0.1$ m. The convergence
area of the condition of Eq.~\ref{eq:convergenceTest} is 
inside the red arc. 200 PC pairs was used to produce the plot.

The final values for the Go-ICP granularity coefficients are:
$\sigma_t=\delta_0$ and $\sigma_r=\theta_0$. Each match uses
an iteration stop criterion given in~\cite{yang2015} and 
the only extra control layer is by monitoring that 
two PCs have enough geometric overlap in the matched configuration. 
The overlap $\lambda\approx 1-\gamma$, but a geometric 
calculation using view cone characteristics is used for the actual test. 
This is because PCs contain churn; trees obscure each other and 
some outliers occur everywhere in the scanning view. If $\lambda$
is not large enough, $\lambda<\lambda_0= 0.4$, the frame pair $(i,j)$ 
will not be used in the iterative improvement of the matches.

\begin{figure}[h!]\centering 
\includegraphics[width=10cm]{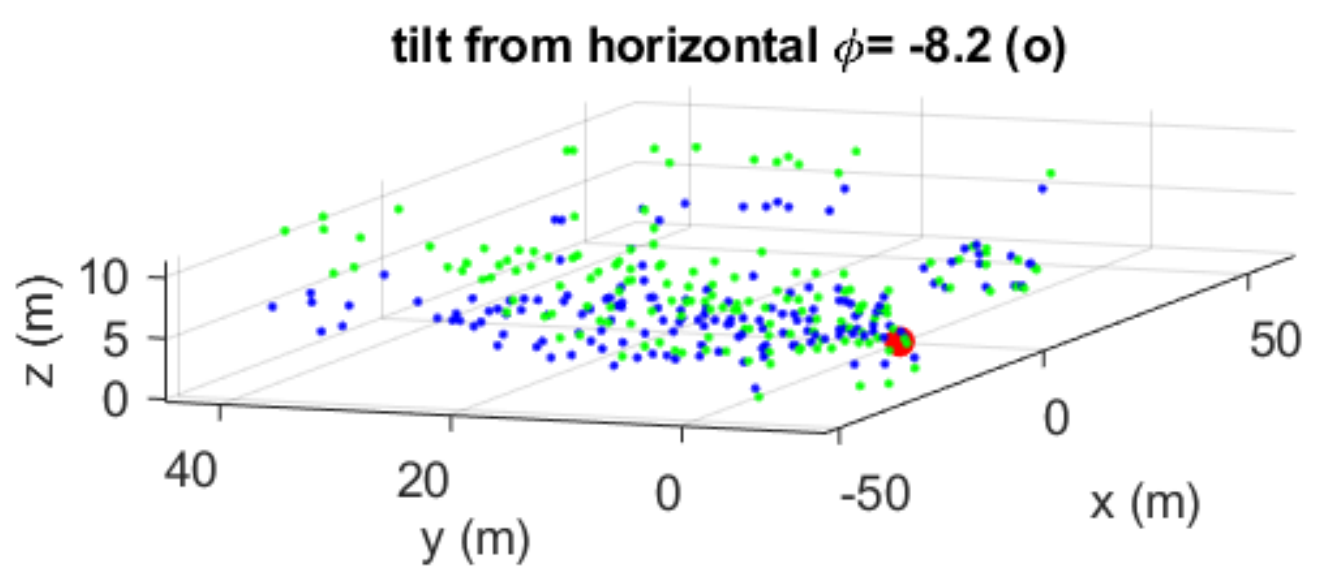}
\caption{A demonstration of the found limit case for
a rather certain ICP match to occur. The angle $\phi=8.2^o$
is the deviation from the vertical axis.} 
\label{fig:pitch}
\end{figure}

The odometry is done in 3D and $\sigma_r$ concerns also  
the roll (and pitch) of the vehicle, even these had 
a negligible effect in point matching. 
This is because the point cloud is relatively flat, see 
Fig.~\ref{fig:pitch}. The Figure depicts also a limit chosen 
$\phi_0= 8.2^o$ for a succesful match. This was found by a numerical 
test producing a similar plot as shown in 
Fig.~\ref{fig:convergenceCondition}. 
The size of $\phi_0$ indicates that the BnB search grid should be 
elongated (it is cubic grid in Go-ICP implementation). 
Very large rolls or pitch movements did not occur, and so 
we limited the BnB search space of rotation to $\pm30^o$ horizontal 
zone and trusted that the hierarchical BnB quickly eliminates the useless
search space. So, the final ICP convergence test is: 
$(\delta/\delta_0)^2 + (\theta/\theta_0)^2 + (\phi/\phi_0)^2\le 1$.

\subsection{Iterative improvement}\label{sec:algorithm}
An initial tree registration and SLAM over sparse PCs is to be 
done at the autonomous vehicle. The aim of the initial SLAM is 
the immediate collision avoidance and basic orientation along the vehicle 
tasks goals. The PCs of selected views 
$V= \{(PC_l,\tau_{l1}\}_{l=1...n}$
will be sent to the cloud environment, where an improvement of the map will occur.

\begin{figure}[h!]\centering 
\includegraphics[width=6cm]{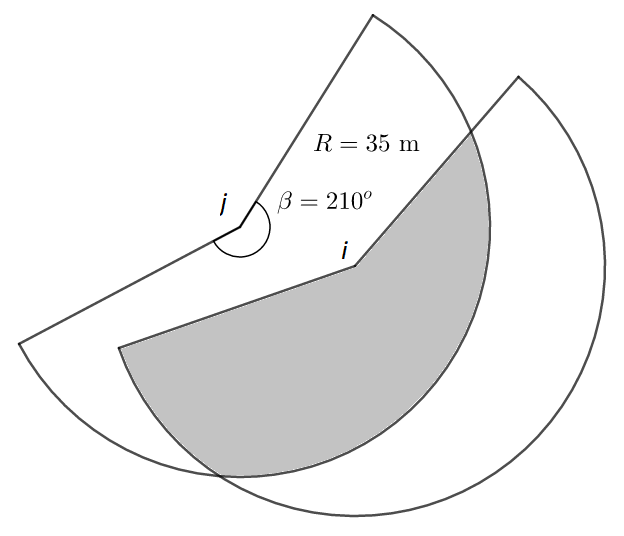}
\caption{Overlap (gray area) can be estimated from the orientation
and position only under an assumption of uniform PC distribution.} 
\label{fig:syntheticOverlap}
\end{figure}

The estimation of the overlap $\lambda$ is depicted in 
Fig.~\ref{fig:syntheticOverlap}. Two vehicle poses 
$\tau_{i1}$ and $\tau_{j1}$ in views $V$ are depicted by their view cones. 
We noticed that the vertical dimension and the corresponding rotations corresponded
very little to the final SLAM map via the match pair selection. Therefore e.g. 
the overlap analysis is done in a projective horizontal plane.

\begin{figure}[h!]\centering 
% \begin{figure*} makes the figure take up the entire width of the page
\includegraphics[width=8cm]{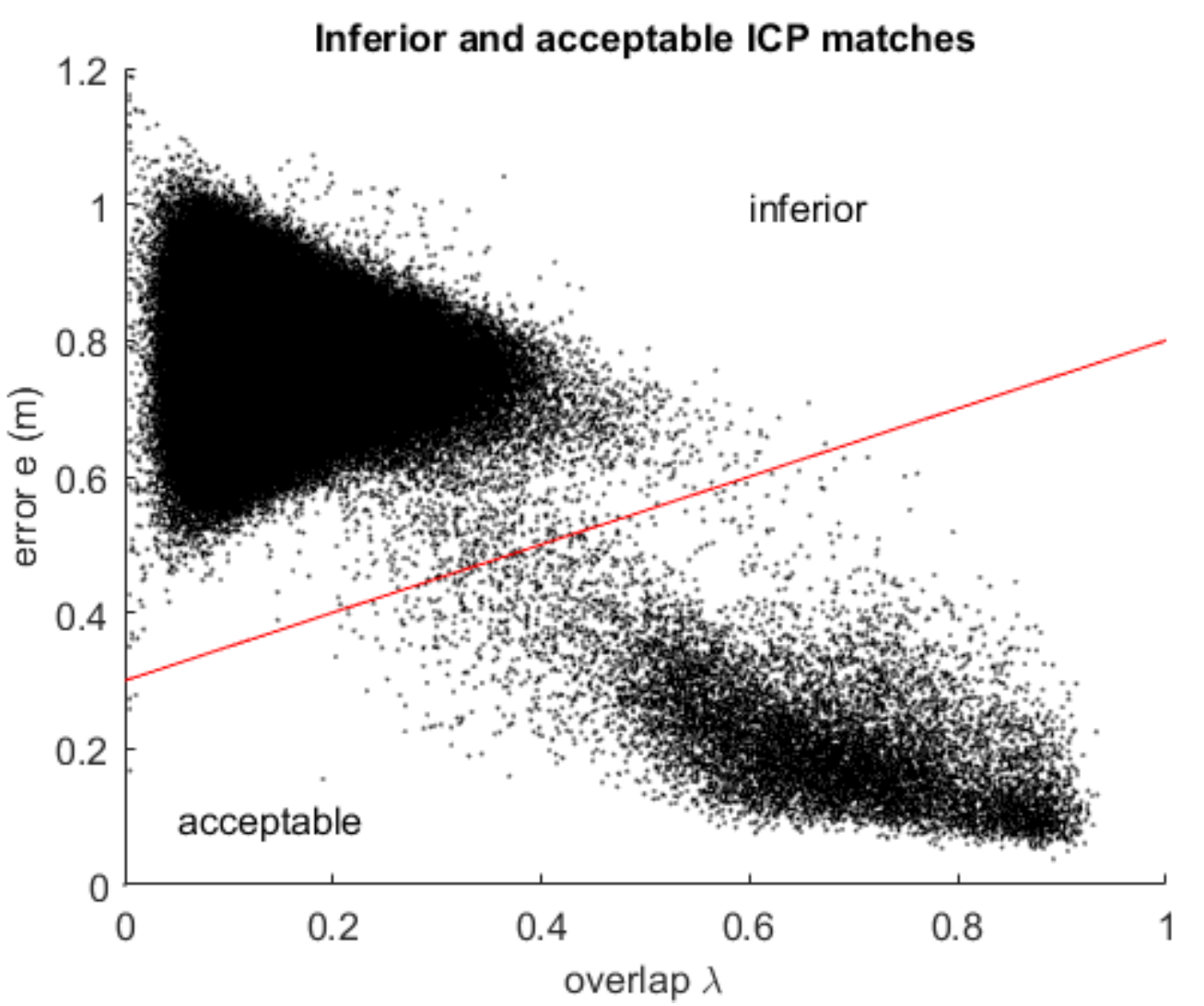}
\caption{Scatter plot over ICP matches. An ICP match attempt either succeeds 
(acceptable) or is inferior (either fails completely or produces small overlap 
and large error.} 
\label{fig:initialDivision}
\end{figure}

The process starts by selecting $m$ potential pairs of frames, which 
will be subjected to improvement. Fig.~\ref{fig:initialDivision} depicts 
a scatter plot based on estimated overlap of views $V$ based on the relative
view cone overlap $\lambda$ and the mean error $e(P_j t_{j1}\match P_it_{i1})$ 
of the match, where non-overlapping parts are excluded. The maximum frame difference 
$\max i-j = 1000$ and 0.2 \% of the inspected 1870000 frames fall
to a promising or acceptable set. The acceptable set was found by looking
overlap ratios over $0.2 < \lambda$ and checking the match error 
$e(P_i t_{i1}\match P_j t_{j1})$ of pairs $(i,j)$. 
% In a real time application this initial search can be done by random tests.

\begin{figure}[h!]\centering 
% \begin{figure*} makes the figure take up the entire width of the page
\includegraphics[width=10cm]{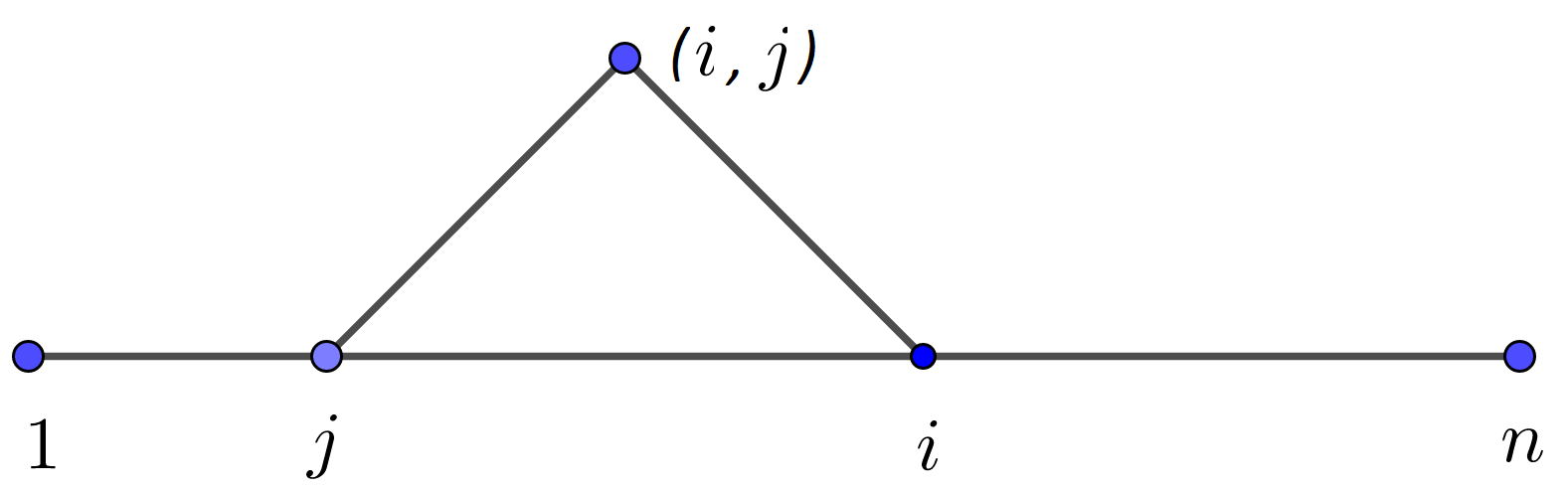}
\includegraphics[width=12cm]{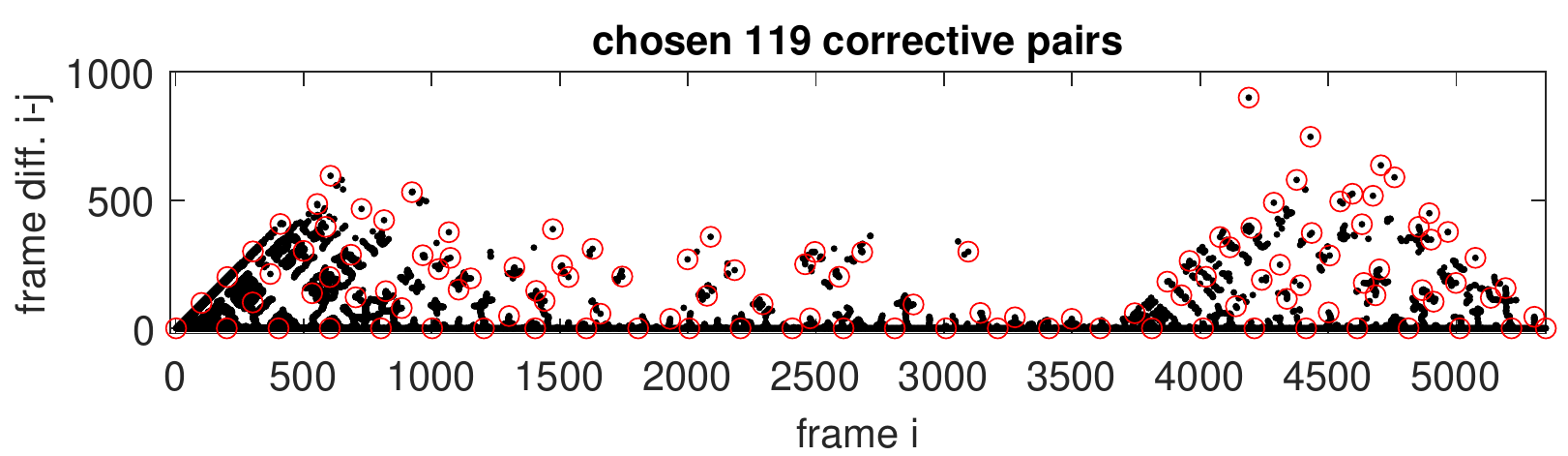}
\caption{
The selection of extra matches. 
Above: A match pairs $(i,j)$ representing a potential match is related 
to frames $i$ and $j$. The original SLAM proceeds along the abscissa 
from 1 to $n$ with $i=j+1$. 
Below: Set $I$ (red circles) of improvement matches with nearly Poisson disk distribution 
are chosen from 2800 potential match pairs of a set $I_2$ (black dots).} 
\label{fig:correctivePairs}
\end{figure}

The preliminary set $I_2\subset [1,n]\subset\mathbb{N}$ depicted by 3500 black dots 
is sampled 
to a subset of $m\approx 100$ nearly Poisson disk distributed pairs depicted by red 
circles in Fig.~\ref{fig:correctivePairs}. The detail at the top of 
Fig.~\ref{fig:correctivePairs} is a schematic about how each black dot $\iota=(i,j)\in I_2$ 
relates to two frames $j$ and $i$.
The nearly Poisson disc sampling was chosen since one can assume an individual 
sample will 
improve the surrounding pairs with equal amount everywhere. A mini-algorithm for
producing a promising sample selection $I$ follows:
\begin{enumerate}
   \item Test recent views $(P_l,\tau_{l1})$ randomly and select pairs $(i,j)$ 
         with $0.2<\lambda$. From those, select ones with the following condition
         fulfilled:
         \begin{equation}
           e_{ij} < 0.3\text{ m}+ \lambda_{ij}\times 0.5\text{ m}
           \label{eq:inequality}
         \end{equation}
         and add $(i,j)\in I_2$. The inequality border is depicted by 
         a red line in Fig.~\ref{fig:initialDivision}. Note that evaluation 
         of values $e_{ij}$ and $\lambda_{ij}$ is relatively cheap, 
         since the former comes from a direct nearest neighbor search and 
         the latter is estimated directly from the parameters of 
         transformations $\tau_{l1}(\theta_l,{\vec\omega}_l,p_l)$ with 
         $l\in\{i,j\}$.
   \item Round $I_2$ to a set of grid points with spacing $\epsilon$.
   A set rounding operator $\{.\}_\epsilon$ is introduced for that purpose: 
   $\{A\}_\epsilon=\{\text{round}(a/\epsilon)\epsilon\;|\, a\in A\}$
   The spacing $\epsilon$ is decreased iteratively by $\epsilon:= 0.8\epsilon$ until 
   the size $|I_1|$ of the rounded set is closest to the intended size: 
   $|I_1|\approx m$. The initial guess is 
   $\epsilon= \sqrt{\text{box area in Fig.~\ref{fig:correctivePairs}}/m}$. With
   the final $\epsilon$: 
   \begin{equation}
       I_1= \{I_2\}_\epsilon
       \label{eq:rounding1}
   \end{equation}
   \item For each occupied grid point $\nu\in I_1$, choose the nearest match from the set $I_2$:
   \begin{equation}
       I= \{\iota\,|\, \iota=\argmin_{\mu\in I_2}\|\mu-\nu\|,\,\nu\in I_1\}
   \end{equation}
\end{enumerate}

The main step of the algorithm applies Eq.~\ref{eq:tauPropagation1} 
and Eq.~\ref{eq:tauPropagation2} randomly until the whole set $I$
is exhausted or a convergence criterion is fulfilled. The simple 
SLAM is tried first. If it delivers a match error $e$ and 
an overlap $\lambda$ which do not fulfill the condition of Eq.~\ref{eq:inequality},
Go-ICP is called instead.

\subsection{Quality criteria of the final map}
Basically, there are two possible convergence criteria, one expressing 
the mean match error $e_J$ over a subset $J\subset I_1$, another one 
quantifying the quality of the final map. A measure useful for possible
applications of tree maps is the tree registration noise $e_C$~\cite{nevalainen2020a}.
The registration noise is root mean square error (RMSE) of the tree cluster
points from the arithmetic mean of the cluster. 

This study focuses on finding the best possible transformations, 
so we use a numerically faster measure, which addresses the sharpness
of the resulting map image. For that purpose, two grid factors
$\epsilon_1=0.2$ m and $\epsilon_2= 10.0$ m are chosen. The first one
counts 1...4 grid points for a tree with a diameter $D=0.1...0.2$ m and
the second one is conveniently larger than the mean distance between nearest
trees $L_0=3.5$ m given in Section~\ref{sec:bnb}. One can define a blur 
ratio $0< \beta <1$:
\begin{equation}
   \beta= \frac{|\{P\}_{\epsilon_1}|\epsilon_1^2}{|\{P\}_{\epsilon_2}|\epsilon_2^2},
   \label{eq:rounding2}
\end{equation}
where $P=\match_{l=1...n} P_l t_{l1}$ is the SLAM map, and 
$\{.\}_\epsilon$ is the set rounding operator originally defined for 
Eq.~\ref{eq:rounding1}. The numerator of the ratio in Eq.~\ref{eq:rounding2} 
approximates the occupied area in the final map $P$ and the denominator 
estimates the overall area of the map.

The blur ratio $\beta$ is used as a target parameter to be minimized in 
the iterative improvement. It is related to the tree registration noise
$e_C$ by having the minimae at the same time, but the absolute value 
of $\beta$ depends on how much undergrowth and small trees are on the site.
Using $\beta$ in governing the improvement process is a novel feature, 
e.g.~\cite{nevalainen2020a} uses the registration noise $e_C$ instead, 
which requires alpha shape~\cite{akkiraju1995} clustering. An tree clusters
are detected by using $r_\alpha= 0.5$ m as the alpha shape radius, and 
ignoring clusters with less than 15 points.

\section{Results}
\begin{figure}[h!]\centering 
\includegraphics[width=10cm]{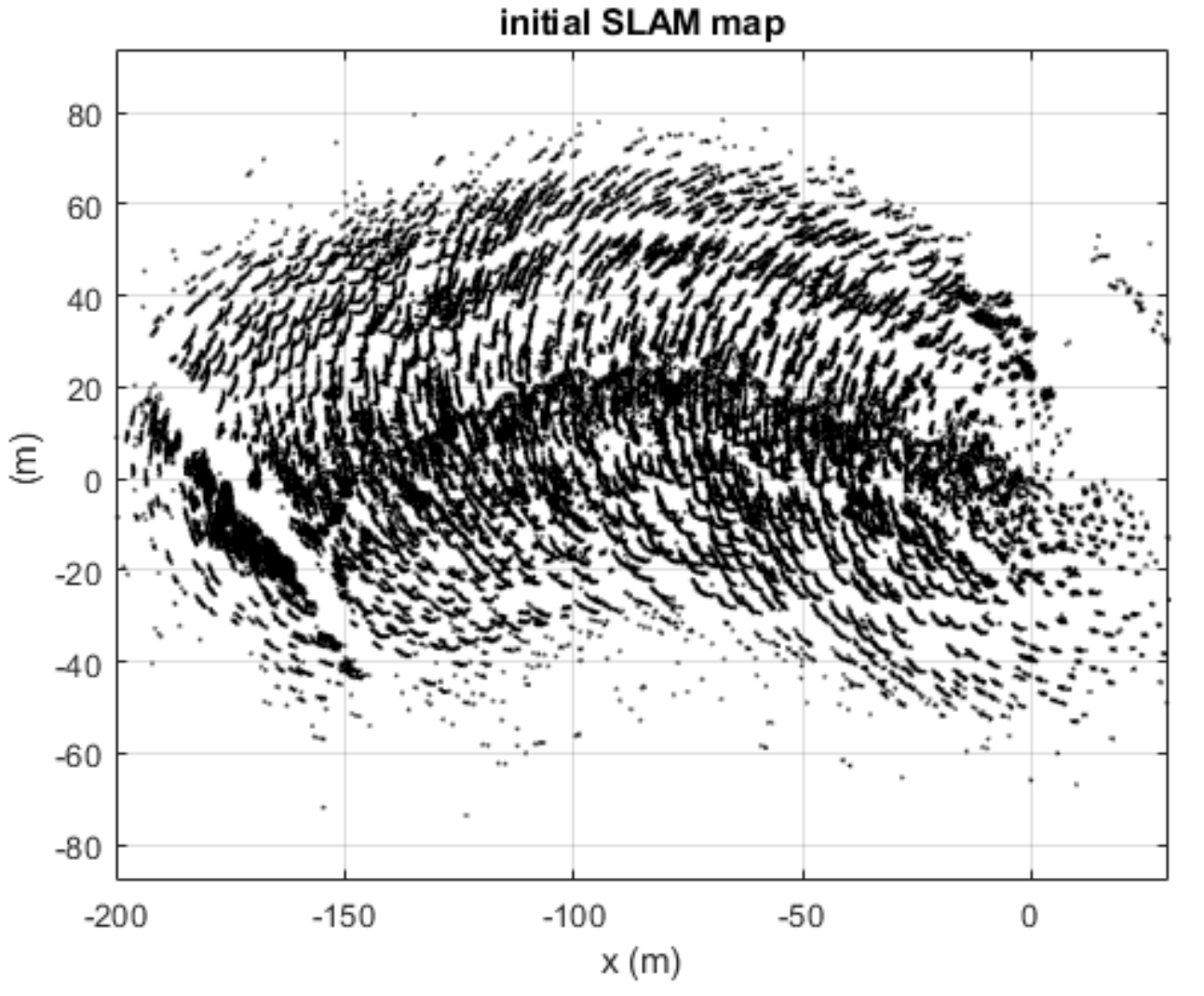}\\
\includegraphics[width=10cm]{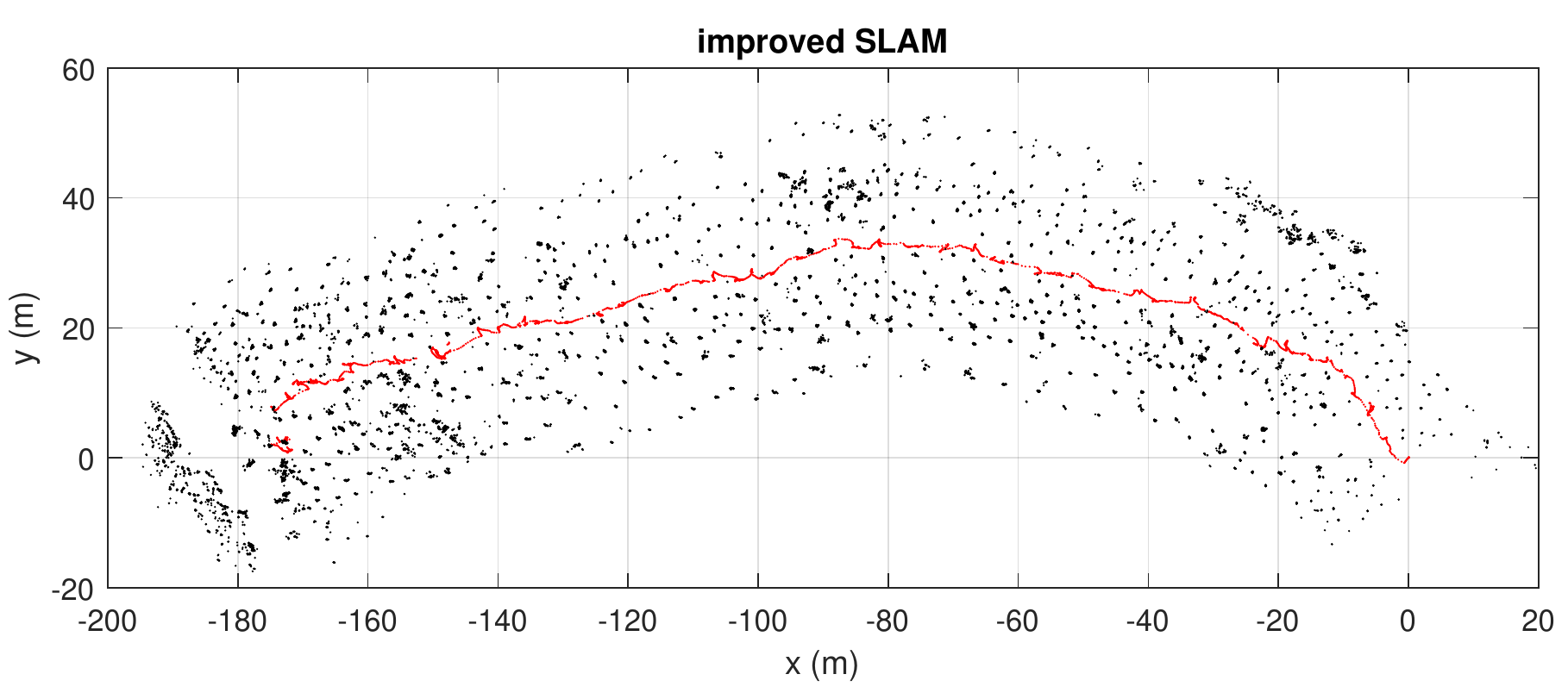}
\caption{Above: the initial SLAM map. 
Below: the final map by the medium-gaps-first strategy.} 
\label{fig:treeMaps}
\end{figure}

The initial (top) and end state (bottom) of the final map 
$P=\cup_{l=1}^n{P_l t_{l1}}$ are depicted 
in Figure~\ref{fig:treeMaps}. Unlike with the ordinary ICP match $.\match .$,
the associations between matching points have not been created but the map 
$P$ is just an unstructured PC. The odometric path is plotted in red. 
The blur ratio $\beta$ of Eq.~\ref{eq:rounding2} moves from the initial $\beta=0.31$ 
of the top map of Figure~\ref{fig:treeMaps} to $\beta=0.022$
of the end map at the bottom. A tree map of practical applications usually 
consists of tree 
cluster centers only, but for the sake of illustration, the registration 
points from all frames have been included. The initial scan on the top 
detail reaches up to 60 m distance while consistent 
tree registrations in the below map are within a 40 m stripe. The simplistic 
SLAM method \textit{pcregistericp}() has a large rotation error, which
is seen from the elongated tree clusters in the top details. The final map 
is the result of the medium-gaps-first strategy. Some blurred
details are undergrowth and thickets.

The average stepwise match error 
$e=\mean_{k=2}^n e(P_{k-1}\match P_kt_{k\,k-1})$ was rather stable 
over the iterative improvement. This is probably because as some 
tree clusters get sharper during the process, some spread out due 
to registration errors.
The order of choosing the corrective pairs of frames 
$\iota=(i,j)$ had a great effect. The strategy of choosing 
the smallest corrective steps first ($i-j$ in ascending order), 
ends to a worse end result $\beta=0.035$ than medium-gaps-first
strategy, which reaches $\beta=0.022$. This result is depicted  
in Figure~\ref{fig:treeMaps}. Three strategies, medium-gaps-first,
smallest-gaps-first and random choice are summarized in 
Table~\ref{table:blur}. The map noise $e_C$ is the RMSE 
of the tree cluster radius.

The map noise value found $e_C=0.15$ m is competitive when compared 
to a similar study~\cite{tang2015} with $e_C\approx 0.2$ m and with 
more dense original scanning. The study areas also compare well, 
our area was $180\times 40\,\text{m}^2$ with approximately 600 trees and 
is gained from over a single open path. 
\begin{table}[h]
    \centering
    \caption{Evaluation times of two ICP methods.}
    \begin{tabular}{@{}lccc@{}}
        \toprule \textbf{Method}\hspace{5em} & \textbf{Blur factor} (1) & \hspace{1.42em}\textbf{RMSE error} (m)\hspace{1.42em} & \textbf{\#Corrections} (m)\\[1pt]
        \midrule \\[-9pt]
        Small-gaps-first
                            &  0.035 & 0.27 & 119 \\[3pt]
        Medium-gaps-first   &  0.022 & 0.15 & 71  \\[3pt]
        Random order        &  0.051 & 0.19 & 119 \\[1pt]
        \bottomrule
    \end{tabular}
    \label{table:blur}
\end{table}

A Python implementation of Go-ICP~\cite{pygoicp} was used. 
The average run time was 0.41 secs over 29 Go-ICP runs triggered.
The average size of the PCs was 131 points making the Go-ICP 
quite fast. 
\begin{table}[h]
    \centering
    \caption{Evaluation times of two ICP methods.}
    \begin{tabular}{@{}lcc@{}}
        \toprule \textbf{Method}\hspace{3em} & \hspace{1em}\textbf{One call} (sec)\hspace{1em} & \textbf{Number of calls} \\
        \midrule\\[-9pt]
          Go-ICP        &  0.41 & 29 \\
          pcregistericp &  0.19 & 80 \\
        \bottomrule
    \end{tabular}
    \label{table:times}
\end{table}

The standard Go-ICP~\cite{yang2015} uses much smaller coefficients 
$\sigma_{r}$ and $\sigma_t$, since it is not specialized to near 
uniform PCs. Also, we found that the tilt $\phi$ from the vertical 
axis of the vehicle is limited to $\phi\in[-\pi/6,\pi/6]$ and 
this further reduces the BnB search space. These advantages are 
summarized in Table~\ref{table:goicp}. The translation granularity 
$\sigma_t$ is computed in~\cite{yang2015} by $\sigma_t=10^{-4}\times N\times L/2$
where $N\approx 130$ is the data point number and $L\approx 180$ m is the largest
included diameter of the PC. This gives an automated value $\sigma_t=1.2$ m.
The translation search was limited to $10\times 10 \times 2\text{ m}^3$ volume in both cases.
\begin{table}[h]
    \centering
     \caption{The BnB search space reduction, when Go-ICP gets adapted 
        to the nearly sparse uniform PCs.}
    \begin{tabular}{@{}lcccc@{}}
        \toprule
           \textbf{PC type}   & \hspace{1em}\textbf{Horiz. zone} ($^o$) \hspace{1em}
                                     & \textbf{Rot. granul.} $\sigma_r$ ($^o$) 
                                             & \hspace{1em}\textbf{Transl. granul.} $\sigma_t$ (m) \hspace{1em}
                                                      & \textbf{BnB size} \\
        \hline General        & 180  &  1.0  & 1.2    & $350\times10^6$ \\
        \hline Sparse uniform &  60  &  3.7  & 1.9    & $1.5\times10^6$\\
        \hline
    \end{tabular}
    \label{table:goicp}
\end{table}

\section{Discussion}
More experiments are needed in deciding a sensible strategy over
the application order of improvement matches. The medium-gaps-first
strategy is just a best found for this particular task, and obviously
there is need for some sort of control, e.g. an end condition
to stop the divergence when the blur ratio $\beta$ does not improve
anymore. A probabilistic way for optimizing
both the selection and ordering of the set $I$ of the frame pairs could
arise by applying e.g. probabilistic data association~\cite{bowman2017}
to Delaunay triangle stars used in~\cite{li2020a}.  

The data~\cite{nevalainen2020b} used is a recording of a forest harvester
operation~\cite{nevalainen2020a}. Although the data allowed in developing
some parts of the pipeline, crucial parts are missing. These are: 
the aforementioned search for the fastest converging sequence of corrections, 
functional memory management of recent scanner views and a process 
converting individual views to a memoized global map~\cite{li2020a}, 
and countering possible systematical errors in relatively simplistic tree 
registration method presented in~\cite{nevalainen2020a}, which was used
to generate the test data~\cite{nevalainen2020b}. 

Since the \textit{pcregistericp}() calls dominate (2800 initial matches 
edge-computed and 80 corrective matches versus 29 Go-ICP calls), 
the combined time stays tolerable and promising for a possible 
full implementation. The 2800 initial matches is to be edge-computed 
at the vehicle, and therefore this process is likely a subject of many 
optimizations concerning the sensors, application specific integrated
circuits (ASIC) and algorithmic developments~\cite{liu2019}. 

The blur ratio $\beta$ of Eq.~\ref{eq:rounding2} is very
close to the dimensionality estimation by box counting~\cite{wu2020}.
The iteration starts with a box counting dimension estimate $d=1.4$ 
and gets stagnated to $d\approx 1$ for a long time while cluster archs 
of individual trees get shorter, see the top detail of Fig.~\ref{fig:treeMaps}. 
Then dimensionality moves to the final $d= 0.3$. PC dimensionality would be 
a better iteration progress indicator in that it does not require specific 
parameters $\epsilon_1$ and $\epsilon_2$. Both indicators $\beta$ and $d$ 
apply in the 2D (projective maps) and 3D cases by a change of the power of
$\epsilon$ in Eq.~\ref{eq:rounding2}.

The search space reduction (230:1) showin in Table~\ref{table:goicp} 
has large but indirect effect to the computation, since the BnB 
process is hierarchical and eliminates large swath of search space
rather soon. The reduction in computation time seems to be in 
the range of 3:1 ... 10:1, when uniformity assumptions are applied.   

From the point of view of lightweight computation at the edge and cloud offloading in remote environments, the method we have proposed in this paper presents some inherent benefits. First, by providing a self-corrective approach there is potential to minimizing the drift in localization for autonomous mobile robots operating over large distances in places with a weak or missing Global Navigation Satellite System performance (GNSS-denied environments). For example, UAVs flying under tree canopy in forests for surveying applications that cannot rely on GNSS sensors are a potential application area. Second, by minimizing the size of the PC used for correcting the odometry process, we can provide cloud offloading or multi-robot collaboration even in environments where connectivity is poor and unreliable and latency does not allow for traditional computational offloading. Therefore, large-scale maps can be built at the cloud or within multi-robot systems in remote environments.

Finally, it is worth mentioning that this method can be extended to multiple domains and application areas. From the perspective of the low computational complexity, this method can extend long-term autonomy in mobile robots by reducing the embedded hardware requirements. This in turn related to lower energy consumption and applicability in smaller platforms. Moreover, if landmarks or anchors are well identified, this can also be leveraged within collaborative multi-robot systems, e.g., with micro-aerial vehicles being deployed from ground units in remote environments~\cite{queralta2020sarmrs}. In forests environments in particular, our adaptive and lightweight self-corrective SLAM approach can be used for either canopy or tree stem registration, but other features that are distributed throughout the operational environments could be exploited as well.

The next two Subsections are devoted to the discussion of alternative details of the Methods section.

\subsection{Alternatives for power coefficients}
The choice of power coefficient $u_l$ seems to have great effect 
on the proposed iterative improvement scheme. Just as there are 
alternatives to the proposed interpolation scheme~\cite{zefrani1998}, 
there are alternatives for the formulation of $u_l$ defined in Eq.~\ref{eq:ulRule}: 
\begin{enumerate}
  \item Cumulative measures like the relative odometric path length 
  $u_l=\sum_{k=j+1}^l \|q_k-q_{k-1}\|/\sum_{k=j+1}^n \|q_k-q_{k-1}\|$ 
  of transformations $\tau_{k1}([omega_k],\theta_k,q_k)$ or
  accumulated match errors: 
  $u_l=\sum_{k=j+1}^l e_k/\sum_{k=j+1}^n e_k$. 
  \item A more sophisticated SE(3) metrics. One candidate is a linear 
  combination of relative rotation and translation~\cite{park1995} $d(\tau_{lj})=\sqrt{a\theta_{lj}^2 + b\|q_{lj}\|^2}$, 
  where $\tau_{lj}=\tau_{lj}(\theta_{lj},\vec{\omega}_{lj},q_{lj})$
  and $0<a,b\in\mathbb{R}$ are free positive constants.
  This leads to: 
  \begin{equation}
    u_l=\frac{d(\tau_{lj})}{d(\tau_{ij})}.
    \label{eq:uSE3}
  \end{equation}
  If the scanner view cone is known, the above measure is very close 
  to the mean squared distance between corresponding spots in 
  the two view cones. 
  \item For paths with a lot of loops, one can find the nearest fit from 
    the skew path $\{\tau_u\,|\tau_{j1}^{1-u}\tau_{i1}^u\}_{0\le u\le 1}$
    shown in Figure~\ref{fig:aSkew}:
\begin{equation}
   u_l=\argmin_u d(\tau_u^{-1}\tau_{lj}), 
   \label{eq:projection}
\end{equation}
where $\tau_u^{-1}\tau_{lj}$ is a transformation from $\tau_{lj}$ to $\tau_u$.
\end{enumerate}

\subsection{Frame elimination}\label{sec:selection}
After the iterative improvement, some frames may show a large 
detrimental contribution to the final map quality. These frames can 
be removed. For this, one has to reshuffle the summation of the map
error $e_C$ to individual frames $l,\,1\le l\le n$:
\begin{equation}
  e_C^2= \mean_{i\in [1,n],p\in {P_i}'}\|p-c_{h_i}\|^2= \sum_{i=1}^n w_i,
\end{equation}
where an inclusion of $p\in P_i'\subset P_i$ occurs only if it 
contributes to some tree in the final map and $w_i$ are the rearranged
summand parts of the mean. The largest values can be removed.
Finding a subset of frames to be removed is a combinatorially 
expensive operation, which should be done only if the application 
specifically requires it. Removing low-quality frames has similarities
with the problem of selecting and ordering the corrective frame pairs;
and both problems resemble feature selection over large feature space 
in general Machine Learning.

\section{Conclusion}\label{sec:conclusion}
This article gives a complete presentation of mathematical details of 
rigid body interpolation and its application to iterative SLAM improvement.
A main motivation was to provide a unified approach to the SLAM accuracy 
improvement. This resulted in an outline of the proposed iterative improvement 
algorithm. The second motivation was to test how much the very reliable 
Go-ICP algorithm can gain advantage from the small and sparse problems.
It seems that Go-ICP is a feasible choice for tree map related SLAM.

Our results suggest that the iterative SLAM improvement using rigid body 
interpolation proposed in this paper
has potential for many applications with sparse PCs, whether point clouds are 
key points, sets of beacons or subsampled PCs. The near uniform distribution makes 
the BnB search grid of Go-ICP coarser, and this and small PC size speeds up 
Go-ICP, which is otherwise known to be a rather slow method. The sensor fusion
with GNSS, inertial mass units and other sensors has been left out to keep 
the presentation simple.

More research is needed especially about an optimal selection of the 
improvement matches before an effort to build a true pipeline from  
autonomous vehicle to a cloud environment can be done. The pipeline would cover 
the edge-computed tree registration and SLAM, transmission of sparse PCs to 
the cloud computing environment and the iterative tree map improvement. 
This may take years, but could be worth of an effort.

%%     ACKNOWLEDGEMENT AND BIBLIOGRAPHY     %%
\begin{acknowledgement}
% This work was supported by the Academy of Finland's AutoSOS project 
% with grant number 328755.
The data was gathered in co-operation with Stora Enso Wood Suply Finland,
Mets\"{a}teho Oy and Aalto University.
The data collection was done under the EFFORTE, Efficient
forestry for sustainable and cost-competitive bio-based industry 
(2016-2019) in WP3---Big data databases and applications.

% The data presented in this study are openly available 
%in Harvard Dataverse at 10.7910/DVN/IO7PZO,~\cite{nevalainen2020b}
\end{acknowledgement}

\bibliographystyle{unsrt}
\bibliography{refs}

\end{document}